# Experience enrichment based task independent reward model

Min Xu

Computational Biology Department

School of Computer Science

Carnegie Mellon University

5000 Forbes Ave, Pittsburgh, PA 15213

Email: mxu1@cs.cmu.edu

**Abstract**

For most reinforcement learning approaches, the learning is performed by maximizing an accumulative reward that is expectedly and manually defined for specific tasks. However, in real world, rewards are emergent phenomena from the complex interactions between agents and environments. In this paper, we propose an implicit generic reward model for reinforcement learning. Unlike those rewards that are manually defined for specific tasks, such implicit reward is task independent. It only comes from the deviation from the agents' previous experiences.

## 1 Introduction

Reinforcement learning [7] focus on designing software agents that learn to take actions in an environment in order to maximize of long term cumulative reward. In recent years, the field of reinforcement learning has been greatly advanced thanks to the revolutions in big data machine learning techniques, especially deep learning [1]. Supervised deep learning enabled the learning of complex rules among large amount of training data. Unsupervised deep learning techniques such as Generative Adversarial Networks (GAN) [2] has been shown to be able to capture the intrinsic distribution on the manifold among images. The use of deep learning for learning value function, policy and model has greatly advanced reinforcement learning field [3]. For example, deep reinforcement learning techniques have been used for playing Atari games and outperformed human experts [5]. Most notably, deep reinforcement learning techniques have also outperformed world champion on playing GO game [6], which has been computationally infeasible through conventional searching techniques due to its huge search space.

Generally a reinforcement learning algorithm aim at maximizing a manually defined long term cumulative reward score that is specific to a particular task. However, in real world, rewards are spontaneous emergent phenomena. Therefore, it is appealing to design general purpose agents to reply on an implicate reward that does not rely on specific tasks, and in the same time, as a side product, can spontaneously produce the intuitive rewards that come from specific tasks. In this paper we propose a reward model that is task independent. Such implicate reward is modeled by the enrichment of experience of the agent.

## 2   Construction of implicate reward model

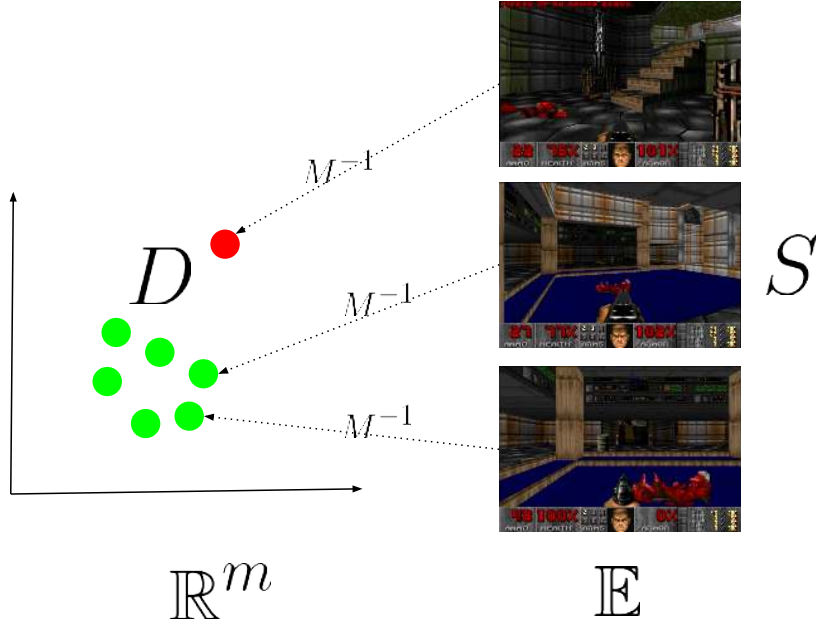

Figure 1: Basic concept of the implicate reward model. Green dots in $\mathbb{R}^m$ correspond with high probability experiences in $\mathbb{E}$. Red dot in $\mathbb{R}^m$ correspond to low probability experience (outliers) in $\mathbb{E}$. The low probability experience will get a high reward. The samples in $\mathbb{E}$ are depicted using the Doom game [e.g. 4].

We define the "experience" of an agent as a subsequence of action-observation pairs. For the operational advantage, the length of the subsequence is fixed. Denote the space of experiences as $\mathbb{E}$. Denote $S \subset \mathbb{E}$ to be a sample of previous experiences collected by the agent through its interactions with environment. To model the enrichment of experiences, we first learn a low dimension representation of $S$ using unsupervised learning, such as Generative Adversarial Networks (GAN) [2]. Denote $\mathbb{R}^m$ as a Euclidean space for reduced representation of $S$. Such unsupervised learning results in a mapping $M : \mathbb{R}^m \to \mathbb{E}$ that maps a simple multidimensional distribution $D$ (such as $m$ dimensional Gaussian distribution) to the distribution that generates $S$. We then construct an inverse mapping $M^{-1} : \mathbb{E} \to \mathbb{R}^m$ through regression using sample pairs $(M(r), r \sim D)$, where $r \in \mathbb{R}^m$ is drawn from $D$. We then construct a probability density estimation $p_{M^{-1}(S)} : \mathbb{R}^m \to [0, +\inf)$ using points in $M^{-1}(S)$. The implicate reward $r(e)$ for any experience $e \in \mathbb{E}$ will be calculated according to $p_{M^{-1}(S)}(M^{-1}(e))$. A lower $p_{M^{-1}(S)}(M^{-1}(e))$ should result in a higher reward $r(e)$. For

example

$$r(e) := \frac{1}{p_{M^{-1}(S)}(M^{-1}(e))} \quad (1)$$

. The basic concept is illustrated in Figure 1.

Most of current reinforcement learning algorithms can be easily adapted to use the above defined implicate reward in place of the explicate manually defined specific reward obtained from specific environment. However, the reward model $r$ needs to be periodically reconstructed using re-sampled $S$. An on-line or on-fly reconstruction approach would be beneficial.

## 3 Discussion

We expect such a purely curious agent using such implicate generic reward would be able to learn to play certain computer games without knowing the beginning, ending, scores from the game. This is because, intuitively, there is a correlation between the richness of the agent's experience and the agent's ability to play a game: in order to obtain an enriched experience, an agent needs to win current level and enter the next level. To mimic people getting richer experience when growing up in the real world, computer games are often designed to introduce new observational (visual and others) experiences in the next level. And the difficulty of a level is often correlated with the complexity of experiences. The pursuit of richer experience is analogous to a free will, the agent would learn how to adapt to environmental constrains and make use of different kinds of resources to survive in order to get a richer experience introduced in the next game level. Even if a computer game does not introduce any significantly new observational experience in the next level, the scores shown on the screen will still be different than previous levels, which may result in a small amount of implicate reward as defined in the method section.